\documentclass[preprint,12pt]{elsarticle}

\usepackage{titlesec}

\titleformat{\paragraph}
  {\bfseries\itshape} 
  {} 
  {0pt} 
  {} 

\titleformat{\subparagraph}
  {\itshape} 
  {} 
  {0pt} 
  {} 
\usepackage{amssymb}
\usepackage{amsmath}
\usepackage{graphicx}%
\usepackage{multirow}%
\usepackage{amsmath,amssymb,amsfonts}%
\usepackage{amsthm}%
\usepackage{mathrsfs}%
\usepackage[title]{appendix}%
\usepackage{xcolor}%
\usepackage{textcomp}%
\usepackage{manyfoot}%
\usepackage{booktabs}%
\usepackage{algorithm}%
\usepackage{algorithmicx}%
\usepackage{algpseudocode}%
\usepackage{listings}%
\usepackage{algpseudocode}
\usepackage[numbers]{natbib}
\usepackage{booktabs}

\begin{document}

\begin{frontmatter}



\title{Adaptive Fusion Network with Temporal-Ranked and Motion-Intensity Dynamic Images for MER}


\author[1]{Thi Bich Phuong Man}
\author[1]{Luu Tu Nguyen}
\author[1]{Vu Tram Anh Khuong}
\author[1]{Thanh Ha Le}
\author[1]{Thi Duyen Ngo\corref{cor1}}

    \affiliation[1]{organization={Faculty of Information Technology, VNU University of Engineering and Technology},
                addressline={144 Xuan Thuy}, 
                state={Ha Noi},
                country={Viet Nam}}

    \cortext[cor1]{Corresponding author: duyennt@vnu.edu.vn}

\begin{abstract}
Micro-expressions (MEs) are subtle, transient facial changes with very low intensity, almost imperceptible to the naked eye, yet they reveal a person’s genuine emotions. They are of great value in lie detection, behavioral analysis, and psychological assessment. However, micro-expression recognition (MER) remains a major challenge due to its short duration, low intensity, and the difficulty of learning discriminative features. Recent studies have exploited deep learning with dynamic images (DIs) and optical flow (OF), but OF is highly sensitive to illumination noise, while traditional DIs – originally designed for action recognition – mainly capture temporal progression without encoding motion intensity. To address these limitations, this paper proposes a novel MER method with two main contributions. First, we propose two complementary representations: Temporal-ranked dynamic image, which emphasizes temporal progression, and Motion-intensity dynamic image, which highlights subtle motions through a frame reordering mechanism incorporating motion intensity. Second, we propose an Adaptive fusion network, which automatically learns to optimally integrate these two representations, thereby enhancing discriminative ME features while suppressing noise. Experiments on three benchmark datasets (CASME-II, SAMM and MMEW) demonstrate the superiority of the proposed method. Specifically, TRDI + MIDI + AFN achieves 93.95\% Accuracy and 89.72\% UF1 on CASME-II, setting a new state-of-the-art benchmark. On SAMM, the method attains 82.47\% Accuracy and 66.52\% UF1, demonstrating more balanced recognition across classes. On MMEW, the model achieves 76.00\% Accuracy, further confirming its generalization ability. The obtained results show that both the input and the proposed architecture play important roles in improving the performance of MER. Moreover, they provide a solid foundation for further research and practical applications in the fields of affective computing, lie detection, and human–computer interaction.
\end{abstract}




\end{frontmatter}



\section{Introduction}\label{introduction}

Emotions play a vital role in daily life, shaping people’s thoughts, decisions, and behaviors \cite{10144523}. Psychological studies have shown that facial expressions are among the most powerful non-verbal channels for conveying emotions, accounting for up to 55\% of emotional communication in social interactions \cite{mehrabian1967decoding}\cite{amsel2019urban}. Facial expressions can be categorized into two types: macro-expressions and micro-expressions. Macro-expressions are more apparent, typically lasting a few seconds, and reflect strong, easily recognizable emotions. In contrast, MEs are unconscious, extremely brief facial movements, usually lasting less than half a second, that reveal genuine emotions which may be suppressed or deliberately concealed. These subtle and fleeting expressions are difficult to perceive with the naked eye but carry valuable information about a person’s internal emotional state. Consequently, MER has become an important research area with applications in lie detection, behavior analysis, and the study of emotional responses in sensitive contexts such as criminal investigations, psychological assessments, and social interactions \cite{Bhushan2015}\cite{yan2013fast}\cite{polikovsky2010detection}.
However, MER has remained a major challenge for many years. The difficulties primarily arise from the intrinsic characteristics of MEs: (1) their extremely short duration makes it challenging to capture and model temporal dynamics; (2) their subtle motion intensity renders the signals highly susceptible to noise and difficult to distinguish from non-relevant facial movements; and (3) the limited availability of annotated datasets restricts the ability of deep learning models to fully exploit their potential. Therefore, achieving high performance in MER requires not only designing an effective input representation but also developing an optimal feature extraction strategy tailored to this type of representation.
In MER, research typically focuses on transforming video sequences into more manageable input forms, primarily categorized into two modalities: static and dynamic\cite{li2022deeplearningmicroexpressionrecognition}. The static modality often relies on the apex frame—the frame that captures the peak intensity of a ME to classify emotions. While this approach reduces the temporal complexity of video analysis, it fails to fully capture the temporal dynamics of MEs, which limits its effectiveness in distinguishing between different emotions.
The dynamic modality addresses the limitations of the static modality by capturing motion changes across video frames and encoding them into image-based representations. Two commonly used inputs in this category are OF and DIs. OF estimates motion vectors (direction and magnitude) between consecutive frames, thereby directly characterizing dynamic information. However, most optical flow–based studies rely on a limited set of representative frames, such as the onset, apex, and offset frames (corresponding to the start, peak, and end of the expression). While this approach can reveal motion changes in MEs, it remains incomplete when restricted to a few representative frames. Moreover, the computational cost increases significantly with the number of optical flow images, and the method is highly sensitive to noise.

Unlike OF, a DI compresses the entire video sequence into a single image using a ranking function\cite{bilen2017actionrecognitiondynamicimage}\cite{li2022deeplearningmicroexpressionrecognition}. This representation not only preserves the temporal order of frames but also reduces computational cost by summarizing dynamic information in a compact form. DIs have demonstrated effectiveness in various action recognition tasks, highlighting their potential for MER. However, when applied to MER, DIs face a critical limitation: they primarily encode the temporal order of frames without explicitly representing motion intensity (MI). Since MEs are characterized by extremely subtle and low-intensity movements, this omission reduces the ability of DIs to capture the most discriminative features. Several extensions, such as active imaging\cite{9207718}, affective-motion imaging \cite{verma2021affectivenetaffectivemotionfeaturelearningfor}, and attention-based models, have been proposed to address this limitation. While these methods achieve modest improvements, their accuracy in MER remains relatively low, typically ranging between 35\% and 60\%. Moreover, approaches like active and affective imaging still rely on the assumption that later frames carry more importance than earlier ones—a premise that may not always hold true in MER scenarios.

Given the limitations of current input modalities—namely, the insufficient ability of static modality to capture temporal kinematic information, the susceptibility of optical flow to noise and high computational cost, and the inadequate design of dynamic images for the MER task—there is a need for a new representation that can simultaneously preserve temporal progression and highlight subtle motion intensity. Such a representation would provide more discriminative features and improve the overall performance of MER.
The choice of feature extraction and emotion recognition method in MER is also a key factor in effectively capturing ME motion. Existing studies are generally divided into two main approaches: hand-crafted methods and deep learning–based methods. For hand-crafted approaches, techniques such as Local Binary Pattern on Three Orthogonal Planes (LBP-TOP), LBP-SIX, Histogram of Oriented Optical Flow (HOOF), and Histogram of Oriented Gradients (HOG) are commonly used to extract ME features\cite{zhao2007dynamic}\cite{10.1007/978-3-319-16865-4_34}\cite{Wang2015EfficientSL}\cite{zhao2007dynamic}\cite{Li2016SpontaneousFM} \cite{Liu2022}. These methods describe facial muscle movements based on pixel motion directions or spatio-temporal texture features. However, they are not powerful enough and remain limited in their ability to capture the subtle and fine-grained dynamic information of MEs.
For deep learning–based approaches, Convolutional neural networks (CNNs) are commonly used for feature extraction and MER. Deep CNN models have been widely applied in MER with the aim of jointly learning both spatial and temporal features. Among them, spatially oriented CNN architectures such as ResNet, VGG, and Inception mainly focus on extracting fine-grained facial features from static frames, especially apex frames\cite{9567872}\cite{10248754}\cite{Zhou2023}. However, a key limitation of this approach is its inability to capture temporal dynamics. To address this, temporal modeling techniques such as Long short-term memory (LSTM) , Gated recurrent unit (GRU), and 3D-CNN have been employed to analyze frame sequences and capture motion information\cite{inproceedings5}\cite{Li2019}. More advanced methods, including 3D-FCNN, CNNCapsNet, and MSCNN, incorporate OF into spatiotemporal deep learning frameworks to enhance performance\cite{sanchezcaballero20203dfcnnrealtimeactionrecognition}\cite{inproceedings1}\cite{10.1007/978-981-96-1071-6_4}. Nevertheless, these models often require large-scale datasets and are computationally expensive. Hybrid architectures, such as Inception-LSTM and 3D-ResNet, combine spatial and temporal feature learning but at the cost of increased complexity\cite{inproceedings10}\cite{3dresnet}. Overall, although deep CNNs have achieved impressive results in many domains, their dependence on large amounts of training data makes them prone to overfitting in MER, where datasets are inherently limited.

Building on the potential yet notable limitations of DIs as input, and considering the challenges faced by deep learning models due to the limited data available for MER, this study proposes a novel MER method. Specifically, we introduce an Adaptive Fusion Network that takes as input two types of DIs: Temporal-ranked dynamic images, which capture temporal progression, and Motion-intensity dynamic images, which emphasize subtle motion intensity. 

The main contributions of this research are as follows:
\begin{itemize}
    \item We propose novel image representations for MER: Temporal-ranked dynamic image and Motion-Intensity dynamic image. These enable the model to capture both temporal variations and the subtle motion intensity of MEs. Specifically:
        \begin{itemize}
            \item Temporal-ranked dynamic image (TRDI): frames containing MEs are rearranged based on the assumption that the frame ranks decrease symmetrically from the apex frame outward. This approach allows for better modeling of the subtle and transient motions of MEs.
            \item Motion-intensity dynamic image (MIDI): frames are weighted by MI, calculated through OF between each frame and the apex frame. The DI is then generated based on the MI coefficient of each frame.
        \end{itemize}
    \item We propose the Adaptive fusion network (AFN), a novel deep learning architecture designed to maximize the complementary information from the two novel image representations. The core idea of AFN is not to treat the two information sources as independent, but to enable them to interact and complement each other through an adaptive fusion mechanism. Specifically:
        \begin{itemize}
            \item Representation fusion block (RFB): proposed to automatically learn fusion weights using attention, enabling the model to flexibly emphasize the more important information source at each spatial location, instead of relying on fixed aggregation or simple pairing strategies as in previous studies.
            \item Multi-scale channel attention block (MSCAB): consists of two main components: 
                (i) Multi-scale feature extraction with kernels of sizes 1×1, 3×3, 5×5, and 7×7, enabling the model to capture both fine local details and global structural patterns, thus overcoming the limitation of a single receptive field; 
                (ii) SE-Block (channel attention), integrated to automatically adjust channel weights, emphasizing informative ME features while suppressing irrelevant noise.

        \end{itemize}

        By combining these two components, AFN not only enhances feature representation capability but also exhibits adaptive dynamic behavior – a crucial factor in processing MEs, which are inherently short-lived and difficult to detect.

\end{itemize}

Experiments were conducted to validate the effectiveness of the proposed method, including: evaluating the novel input representations, assessing the performance of the AFN architecture, and analyzing the combined effectiveness of both proposed components.

\section{Proposed method}\label{s: proposed_method}

\begin{figure}
    \centering    \centerline{\includegraphics[width=1\linewidth]{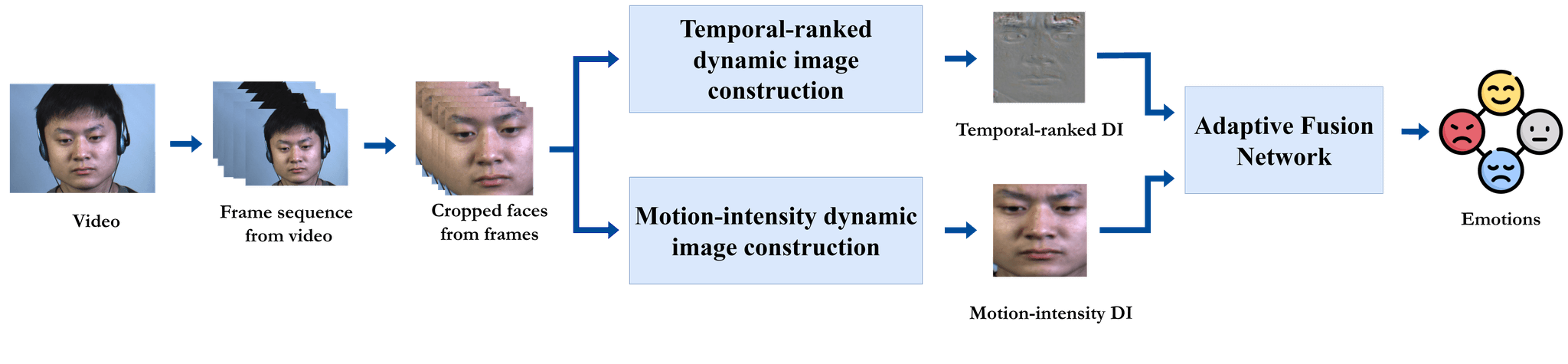}}
    \caption{Pipeline for MER with AFN via TRDI and MIDI}
    \label{fig:pipeline}
\end{figure}

To overcome the limitations of traditional dynamic imaging, this paper proposes an improved MER method by integrating an AFN with two new representations: TRDI and MIDI. Unlike conventional dynamic imaging, which is mainly designed for action recognition, TRDI rearranges frames based on temporal ranking to emphasize subtle and transient changes, while MIDI captures the local motion intensity between frames. These two complementary representations provide richer and more discriminative information for the MER task. Furthermore, the AFN model is designed to effectively fuse the features extracted from TRDI and MIDI, thereby improving the overall recognition performance. 

Figure \ref{fig:pipeline} illustrates the workflow of the proposed method, where TRDI and MIDI refine the input representation, while AFN performs multi-stream feature fusion and optimizes the MER process. The following subsections will detail each component.

\subsection{\textbf{Dynamic image construction}}

Dynamic image was developed by Bilen et al. \cite{bilen2017actionrecognitiondynamicimage} based on the concepts of ranking function and rank pooling proposed by Fernando et al.\cite{fernando2015modeling}\cite{fernando2016rank}. The main idea is to represent a video by a single feature vector through learning a ranking function over its frames.

Specifically, for a sequence of frames $I_1, I_2, ..., I_T$, Fernando defined the ranking function in Equation \ref{equa1}:  

\begin{equation}
S(t|d) = \langle d, V_t \rangle, \quad V_t = \frac{1}{t} \sum_{\tau=1}^t \psi(I_\tau)
\label{equa1}
\end{equation}

where $d \in \mathbb{R}^n$ is a learnable parameter vector, and $\psi(I_\tau)$ denotes the feature vector of frame $\tau$. 
The objective is to ensure that later frames receive higher scores in Equation \ref{equa2}:  

\begin{equation}
q > t \implies S(q|d) > S(t|d).
\label{equa2}
\end{equation}

The optimal vector $d^*$ is estimated by Equation \ref{equa3} \ref{equa4}:  

\begin{equation}
d^* = \rho(I_1, I_2, ..., I_T ; \psi) = \arg \min E(d),
\label{equa3}
\end{equation}

\begin{equation}
E(d) = \frac{\lambda}{2} \|d\|^2 + \frac{2}{T(T-1)} \sum_{q>t} \max \{0, 1 - S(q|d) + S(t|d)\}.
\label{equa4}
\end{equation}

This process is referred to as \textit{rank pooling}, and the resulting $d^*$ can be interpreted as an RGB image representing the whole video.  

Bilen et al. later proposed \textit{Approximate Rank Pooling (ARP)} to reduce computational cost. 
ARP approximates $d^*$ as a weighted sum of frame features in Equation \ref{equa5}:  

\begin{equation}
d^* = \sum_{t=1}^T \alpha_t \psi(I_t), \quad \alpha_t = 2t - T - 1.
\label{equa5}
\end{equation}

In this study, we adopt \textit{ARP} to generate dynamic images from ME frame sequences, which are then used as input for the recognition process.

\subsection{\textbf{Temporal-ranked dynamic image}}
In traditional DI construction, the basic assumption is that frames appearing later in the sequence usually carry more important information, so this method tends to assign higher weights to the ending frames of the video. This assumption is suitable for action recognition tasks, where motion often accumulates towards the end of a segment. However, when applied to MER, empirical observations show that this assumption no longer holds.

\begin{figure}[h]
    \centering    \centerline{\includegraphics[width=0.8\linewidth]{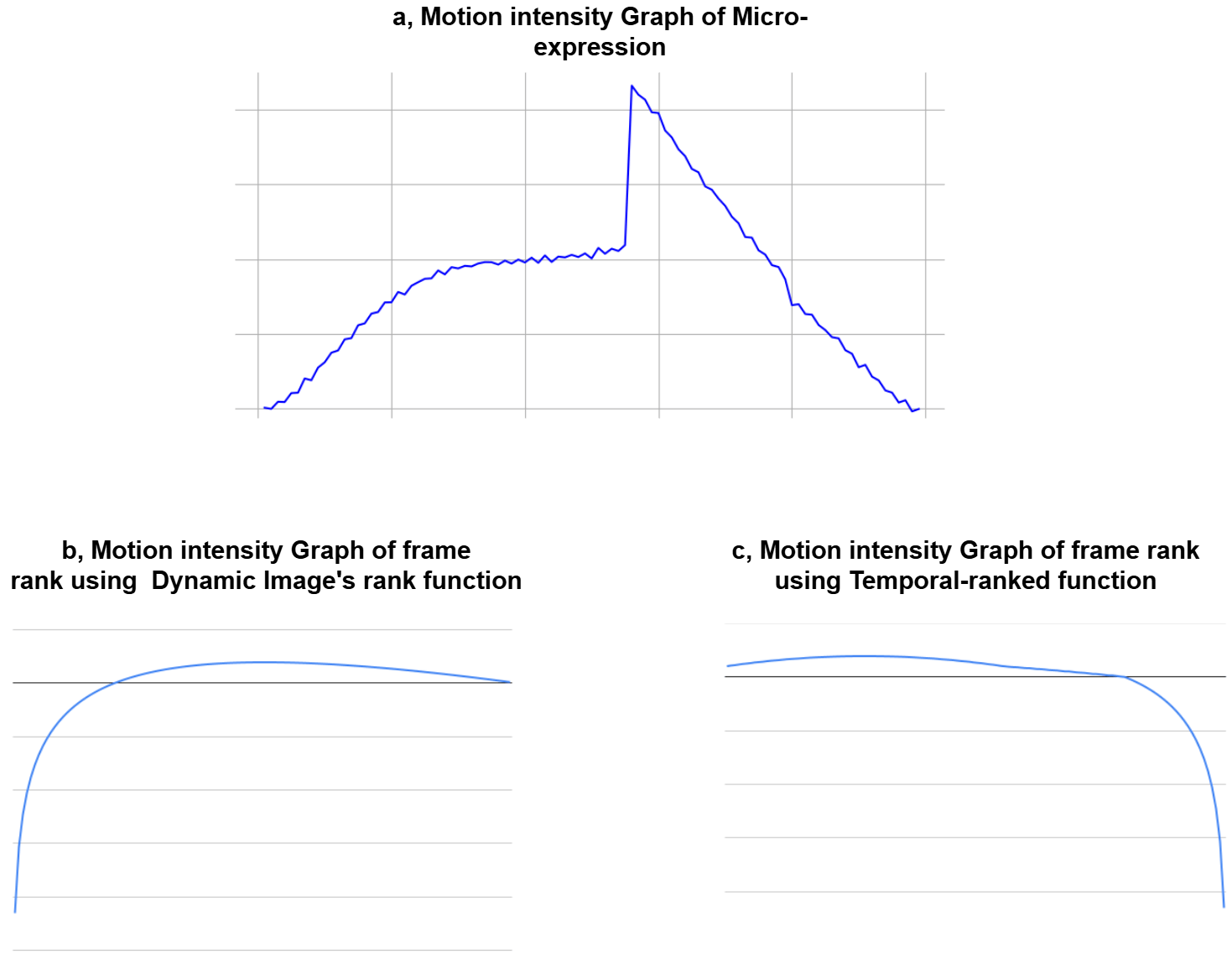}}
    \caption{Comparison of ME motion intensity with DI and Temporal-ranked function}
    \label{fig:graph}
\end{figure}

The distinctive characteristic of MEs is that their motion intensity is very small, short-lived, and mainly concentrated around the apex frame. If depicted graphically, the MI trajectory often resembles a bell shape: gradually increasing from the onset, reaching a peak at the apex, and then gradually decreasing towards the offset (Figure \ref{fig:graph}a). This indicates that the frames around the apex actually contain the most important information for recognition, rather than the final frames as assumed in traditional DIs.

Based on this observation, we propose a Temporal-ranked method, in which DI construction is performed according to temporal ranking. Specifically, instead of summing or averaging the entire frame sequence as in the traditional approach, the frames in the TRDI are rearranged according to a temporal ranking, which typically decreases symmetrically from the central frame (apex) towards both the onset and the offset. This idea reflects the assumption that the central frame (apex) carries the most ME information, while frames further away from the center are less important.

During the construction process, each frame is assigned a weight based on its temporal ranking. As a result, the final TRDI emphasizes the most informative frames around the apex, while reducing noise from less relevant frames. This approach ensures that the representation is not only consistent with the bell-shaped nature of MEs, but also more optimal for the recognition task.

\textbf{Algorithm \ref{alg:redistribute_coefficients}} details the Temporal-ranked mechanism:
\begin{itemize}
    \item First, the motion coefficients are sorted in descending order to prioritize the values that represent the strongest motion.
    \item Next, a new coefficient array is initialized with all values set to 0, retaining only the important coefficients to reduce noise from irrelevant data. The largest weight is assigned directly to the apex frame, reflecting the critical importance of this moment.
    \item Finally, the remaining coefficients are distributed symmetrically around the apex, modeling the propagation of motion from the peak intensity towards both the onset and offset frames.
\end{itemize}

\begin{figure}
    \centering    \centerline{\includegraphics[scale=0.6]{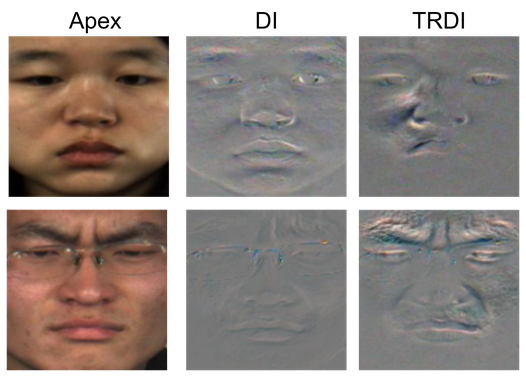}}
    \caption{Comparison between the original DI and the proposed TRDI.}
    \label{fig:di_visual}
\end{figure}

The result is a symmetrical bell-shaped weight distribution, consistent with the empirical characteristics of MEs (Figure \ref{fig:graph}b, \ref{fig:graph}c). Comparisons show that this representation better emphasizes meaningful frames and smooths the transitions, resulting in a more informative and less noisy DI. As shown in Figure \ref{fig:di_visual}, we compare apex frame, the original DI and the TRDI. It can be seen that TRDI highlights the details around the eye and forehead regions more clearly, while the original DI still contains a lot of irrelevant noise.

This makes the Temporal-ranked method particularly effective: it not only overcomes the limitations of traditional DIs but also fully leverages the natural characteristics of MEs. By prioritizing frames around the apex, the deep learning model can directly exploit the most informative signals, thereby enhancing discriminative power and significantly improving MER performance.

\begin{algorithm}[H]
\caption{Temporal-ranked method for dynamic image}
\label{alg:redistribute_coefficients}
\footnotesize
\begin{algorithmic}[1]
\Require A coefficient array $\texttt{coefficients}$ of length $n$, an apex frame index $\texttt{apex\_frame}$.
\Ensure A new coefficient array $\texttt{new\_coefficients}$.

\State $\texttt{sorted\_coefficients} \gets \texttt{SortDescending(coefficients)}$
\State $\texttt{new\_coefficients} \gets [0, 0, \dots, 0]$ \Comment{Initialize with zeros}
\State $\texttt{new\_coefficients[apex\_frame]} \gets \texttt{sorted\_coefficients[0]}$

\For{$i = 1$ to $\lfloor n/2 \rfloor$}
    \If{$\texttt{apex\_frame} - i \geq 0$}
        \State $\texttt{new\_coefficients[apex\_frame - i]} \gets \texttt{sorted\_coefficients[i]}$
    \EndIf
    \If{$\texttt{apex\_frame} + i < n$}
        \State $\texttt{new\_coefficients[apex\_frame + i]} \gets \texttt{sorted\_coefficients[2i - 1]}$
    \EndIf
\EndFor

\State \Return $\texttt{new\_coefficients}$
\end{algorithmic}
\end{algorithm}

\subsection{\textbf{Motion-intensity DI construction}}
One of the major challenges in MER is how to extract the subtle and transient features of facial motion. Although TRDI is capable of emphasizing the importance of frames based on their temporal positions, it does not capture the differences in motion intensity between frames. In practice, not all frames contribute equally: some frames exhibit distinct motion associated with the ME apex, while many others are mostly neutral or contain substantial noise. Based on this observation, we propose the MIDI method, which incorporates motion intensity into the DI generation process to emphasize the most discriminative frames.  

The core idea of MIDI is based on the assumption that MEs are typically characterized by localized and brief facial muscle activations, with MI varying over time. Therefore, frames with larger motion magnitudes should be assigned higher weights during the synthesis process, while frames with weaker or irrelevant motion should be down-weighted. Incorporating motion magnitudes into the DI generation helps to highlight important signals while minimizing the impact of background noise or irrelevant variations, thus providing a clean yet informative representation for the recognition model.  

To compute MIDI, we estimate the OF between each frame and the apex, assuming that the apex contains the maximum motion of the ME. The DIS optical flow algorithm is employed due to its ability to balance computational efficiency with sensitivity in detecting small displacements. For each frame pair, we obtain a two-dimensional OF field $(dx, dy)$, from which we calculate the average motion amplitude over the entire image.

For each frame $i$, the motion magnitude is computed as follows Equation \ref{equa6}:  

\begin{equation}
m_i = \frac{1}{HW} \sum_{x=1}^{W} \sum_{y=1}^{H} \sqrt{dx(x,y)^2 + dy(x,y)^2}.
\label{equa6}
\end{equation}

To ensure that the apex frame always receives the highest weight, the magnitude values are normalized by inverting them with respect to the maximum value follow in Equation \ref{equa7}:

\begin{equation}
   w_i = \max_j (m_j) - m_i. 
   \label{equa7}
\end{equation}

The final MIDI dynamic image is constructed as a weighted sum of all frames, as shown in Equation \ref{equa8}:

\begin{equation}
    DI_{MIDI} = \sum_{i=1}^{T} w_i \cdot f_i. 
    \label{equa8}
\end{equation}

\begin{figure}[t]
    \centering
    \includegraphics[width=0.8\linewidth, keepaspectratio]{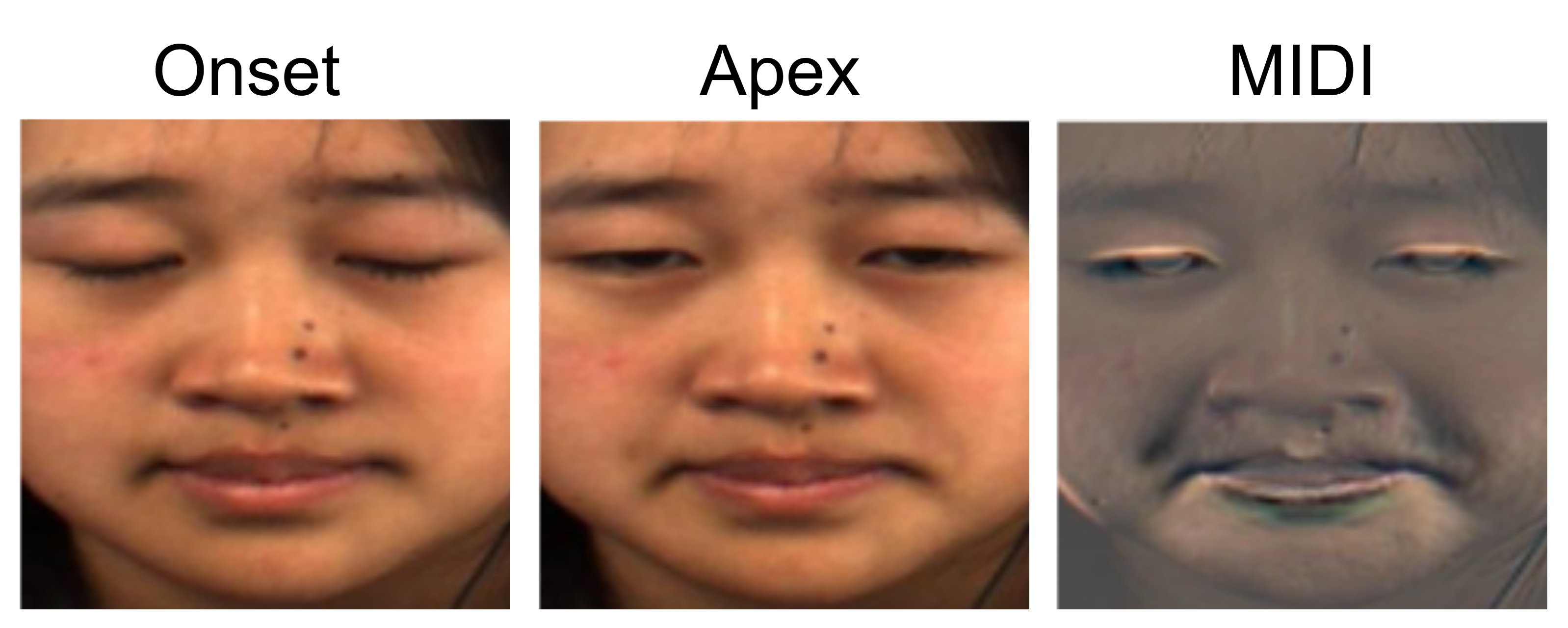}
    \caption{Comparison between Onset frame, Apex frame and the proposed MIDI.}
    \label{fig:midi_visual}
\end{figure}

This approach incorporates motion intensity directly into the DI construction process, allowing the model to focus more on frames that contain key expressive information while reducing the influence of neutral or noisy frames. As shown in Figure \ref{fig:midi_visual}, the MIDI compared with the apex frame and onset frame reveals more clearly the subtle motion cues on the face. This indicates that MIDI not only inherits the critical information from the apex frame but also highlights fine-grained expressive details, making it easier for the model to extract discriminative features. Consequently, MIDI becomes an effective representation that captures the spatio-temporal dynamics of MEs, and is particularly useful in distinguishing extremely subtle expressions where critical signals are often obscured by irrelevant variations.

\subsection{\textbf{Adative fusion network}}
\begin{figure*}
    \centering    \centerline{\includegraphics[width=1\linewidth]{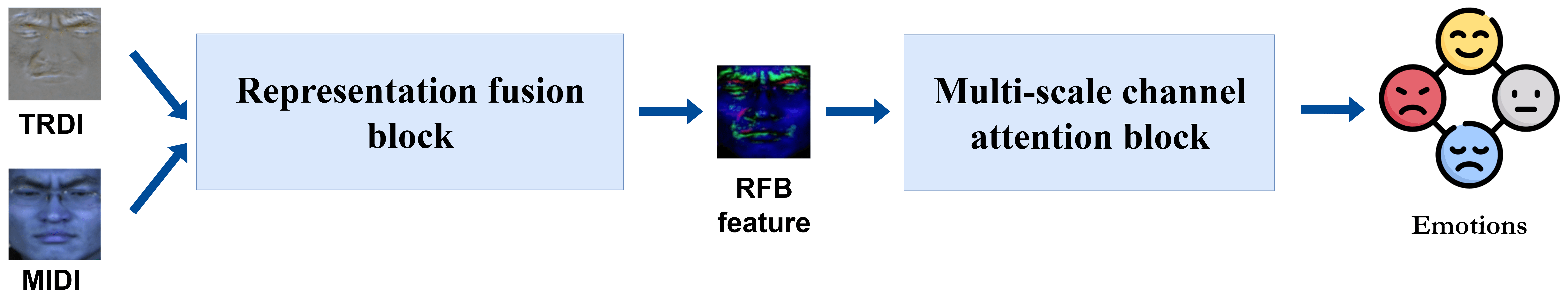}}
    \caption{The architecture of AFN}
    \label{fig:AFN}
\end{figure*}

In MER, DI-based methods face inherent limitations. Using only one type of DI often results in information imbalance, while static fusion methods such as addition or averaging are inflexible because they apply a fixed fusion rule to all instances, regardless of the individual characteristics of each ME. To address these issues, we propose the AFN as shown in Figure \ref{fig:AFN}, an architecture designed to intelligently handle information fusion and feature extraction. 

AFN achieves this through two core components: first, the Representation fusion block employs an attention mechanism to adaptively adjust the weights of each type of representation according to the context, producing an optimal fused representation. Next, the Multi-scale channel attention block leverages this representation by extracting multi-level features and rebalancing the importance of information channels. Through this combination, AFN not only generates a more informative input but also ensures that the most discriminative features are effectively exploited, thereby significantly enhancing recognition performance.

\subsubsection{Representation fusion block}
\begin{figure}
    \centering    \centerline{\includegraphics[width=1\linewidth]{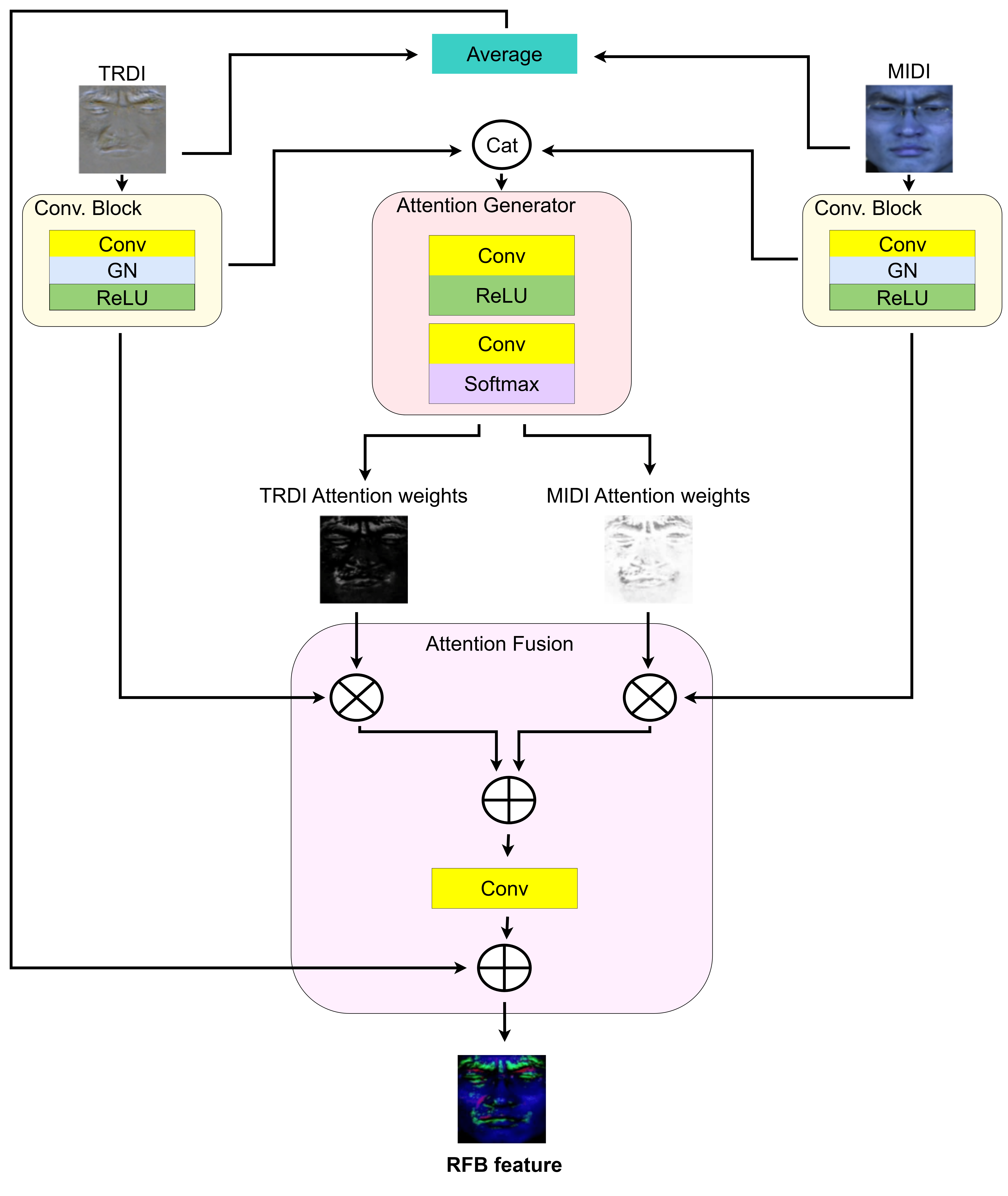}}
    \caption{The architecture of Representation fusion block}
    \label{fig:difusionmodule}
\end{figure}

This paper introduces the RFB as shown in Figure \ref{fig:difusionmodule}, a core component designed to address the challenge of adaptively and efficiently fusing information from multiple DIs. This module is particularly well-suited for integrating our two main input formats: the TRDI and the MIDI. Instead of relying on static fusion methods such as concatenation or pooling, the module employs a spatial attention mechanism to dynamically assign weights to each information source at every spatial location in the image.

Specifically, the module begins by extracting high-level features from each input image (coef\_img and motion\_img) through the coef\_encoder and motion\_encoder branches. Each branch consists of a convolutional layer to capture local spatial patterns, followed by GroupNorm and ReLU to ensure training stability and enable nonlinear representation learning. The extracted features from these two branches are then concatenated and passed through a lightweight neural network (attention\_generator). This network, composed of 1×1 convolutional layers, learns to generate spatial weight maps (attention\_weights). These weights, normalized via a softmax function, determine the contribution of each feature source (coef\_features and motion\_features) at every spatial location.

Finally, adaptive fusion is performed by element-wise multiplying the features from each source with their corresponding attention weights, followed by summation to produce fused\_features. A 1×1 convolutional layer (final\_conv) is then applied to adjust the output channel dimension. In addition, a skip connection is introduced by adding the average of the two original input images. This skip connection not only facilitates gradient propagation through the deep network but also ensures that essential information from the original inputs is preserved, thereby improving training stability.

Through the combination of these mechanisms, the RFB produces an optimal, informative, and context-adaptive fused representation, which serves as a rich input for the subsequent stages of the MER model.

\subsubsection{Multi-scale channel attention block}
\begin{figure}
    \centering    \centerline{\includegraphics[width=1\linewidth]{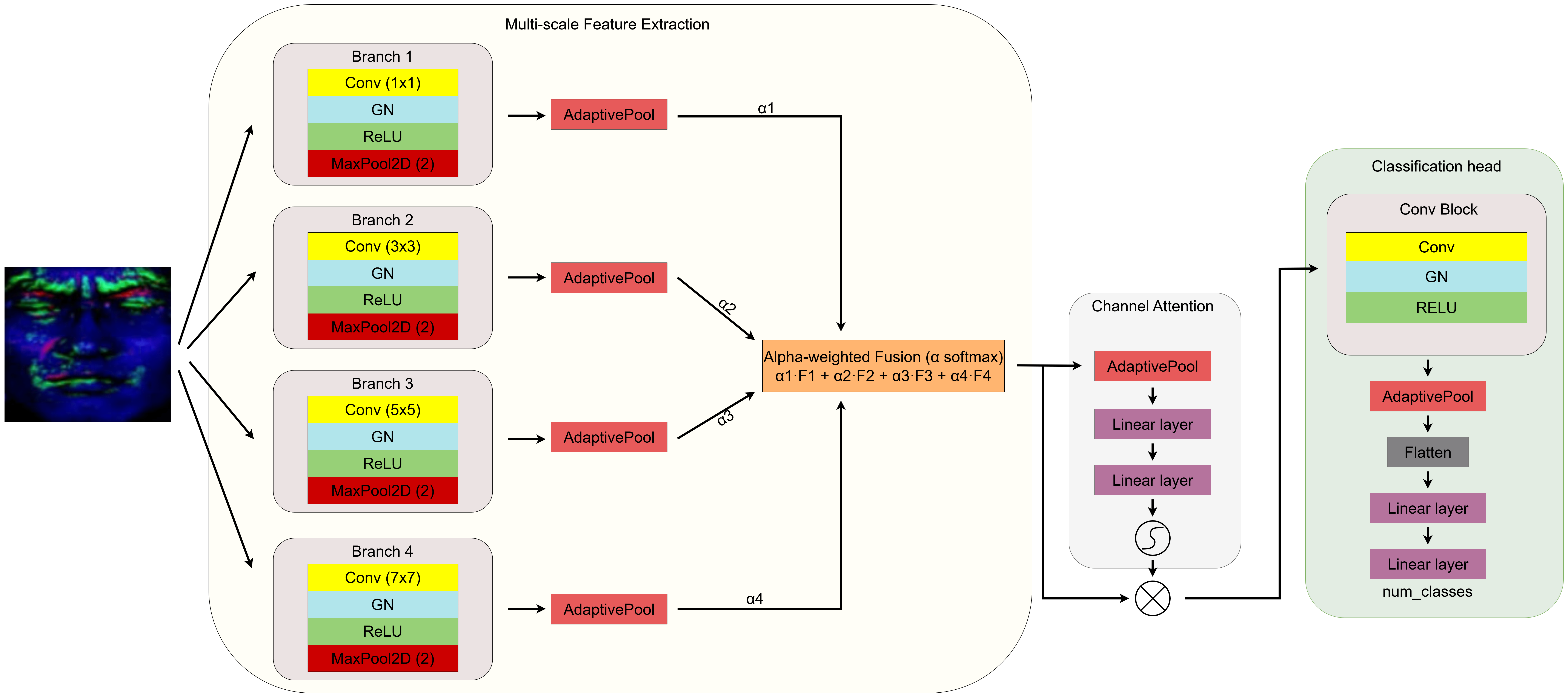}}
    \caption{The architecture of MSCAB}
    \label{fig:MSCAB}
\end{figure}

MSCAB, as shown in Figure \ref{fig:MSCAB} is the second core component of AFN, responsible for extracting discriminative features from the adaptively fused DIs and optimizing their representations for the MER task. While the RFB ensures that the input image is both contextual and informative, MSCAB further enhances this representation by performing deep feature extraction, multi-level abstraction, and channel rebalancing.
The module begins with a backbone feature extractor that processes the fused image to capture hierarchical features, ranging from low-level spatial textures to high-level semantic motion patterns. A lightweight yet powerful convolutional neural network is employed to balance computational efficiency with representational capacity, ensuring that subtle motion cues are preserved.
To further refine the extracted features, MSCAB incorporates a channel reweighting mechanism inspired by the attention principle. This mechanism adaptively emphasizes the most informative channels while suppressing redundant or noisy ones. This is particularly important in MER, where discriminative cues are extremely subtle and sparse. By dynamically adjusting the importance of different feature channels, the model ensures that the core ME signals dominate the final representation.
Finally, the refined features are aggregated and passed through a classification head, which maps them into the ME label space. This head typically consists of a global pooling layer to condense spatial information, followed by fully connected layers to perform classification.
With this design, MSCAB not only maximizes the utility of the fused DIs but also directly addresses the core challenge of MER: distinguishing subtle, low-intensity, and transient expressions. By complementing the adaptive fusion stage, MSCAB ensures that the most discriminative patterns are effectively captured and leveraged for accurate recognition.

\section{Experiment and Results}\label{s:experiment_and_results}

\subsection{Experiment}
To evaluate the effectiveness of the proposed method, a series of experiments were conducted to analyze the contributions of each component individually as well as in combination. The experiments were designed with three main objectives:

\begin{itemize}
    \item First, we assess the effectiveness of two newly proposed dynamic image representations, TRDI and MIDI, specifically designed for the MER task. Visual and quantitative evaluations demonstrate their superiority in capturing and representing subtle MEs.
    \item Second, we examine the performance of the proposed AFN architecture, which is tailored for feature extraction in MER and addresses data limitations. Comparisons with pre-trained DCNN models show that AFN achieves more meaningful feature representations.
    \item Finally, we evaluate the overall performance of the MER framework by integrating TRDI, MIDI, and AFN. The combination of enhanced motion representation from TRDI and MIDI with optimized feature extraction from AFN results in significant performance improvement, confirming the complementary nature of these components.
\end{itemize}

\subsubsection{Dataset} 
This study utilizes three widely used MER benchmarks: CASME-II \cite{Yan2014-za}, SAMM \cite{7492264}, and MMEW \cite{9382112}.

\textit{CASME-II:} Consists of 247 videos from 26 participants recorded at 200 fps. Videos were annotated using FACS, and only emotion labels with sufficient samples were retained, resulting in five categories: surprise, happiness, disgust, other, and repression.

\textit{SAMM:} Contains 159 videos from 32 participants across 13 ethnicities, recorded at 200 fps with high spatial resolution. Five emotion categories were annotated: anger, contempt, happiness, surprise, and other.

\textit{MMEW:} Provides both macro- and micro-expressions from the same participants, with 300 micro-expression and 900 macro-expression samples at 1920×1080 resolution and 90 fps. Four emotion categories were used for experiments: happiness, surprise, disgust, and others.

\textbf{Data augmentation:} To improve generalization, images were horizontally flipped and rotated at small (±5°) and larger (±10°) angles. All images were resized to 224×224, converted to three-channel grayscale, and normalized. Each original image was transformed into multiple variants to expand the dataset and reduce overfitting.

\subsubsection{Experiment settings}

All experiments follow the Leave-One-Subject-Out (LOSO) protocol. In each iteration, samples from one subject are used for testing, while the remaining subjects are used for training. Final performance is reported as the mean accuracy over all subjects. Training configuration includes Adam optimizer with a learning rate of 0.001, weight decay of 0.0001, batch size of 16, 50 epochs, and cross-entropy loss. Two main experiments are conducted:

Experiment 1: Effectiveness of the proposed dynamic image inputs.  
This experiment investigates the contribution of the two proposed dynamic image representations, TRDI and MIDI, for MER. These inputs are designed to capture subtle facial movements and are compared against other input types, including Apex frames and original DIs. Both MSCAB (a component of AFN) and ResNet are used as backbone architectures to evaluate and demonstrate the superior performance of TRDI and MIDI in representing micro-expressions.

Experiment 2: Effectiveness of the complete MER framework.
This experiment evaluates the overall performance of the proposed method, AFN. TRDI and MIDI are used as the primary inputs. Additionally, a combination of MIDI with the original dynamic image is evaluated to investigate whether integrating complementary motion information can further improve performance. To highlight the advantages of the proposed architecture, AFN is tested both with its original MSCAB component and with ResNet as a replacement backbone, allowing a direct comparison of feature extraction and overall MER performance.

This experimental setup ensures reliability, reproducibility, and robustness of the proposed method.

\subsection{Results}

\begin{figure}
    \centering    \centerline{\includegraphics[width=1.3\linewidth]{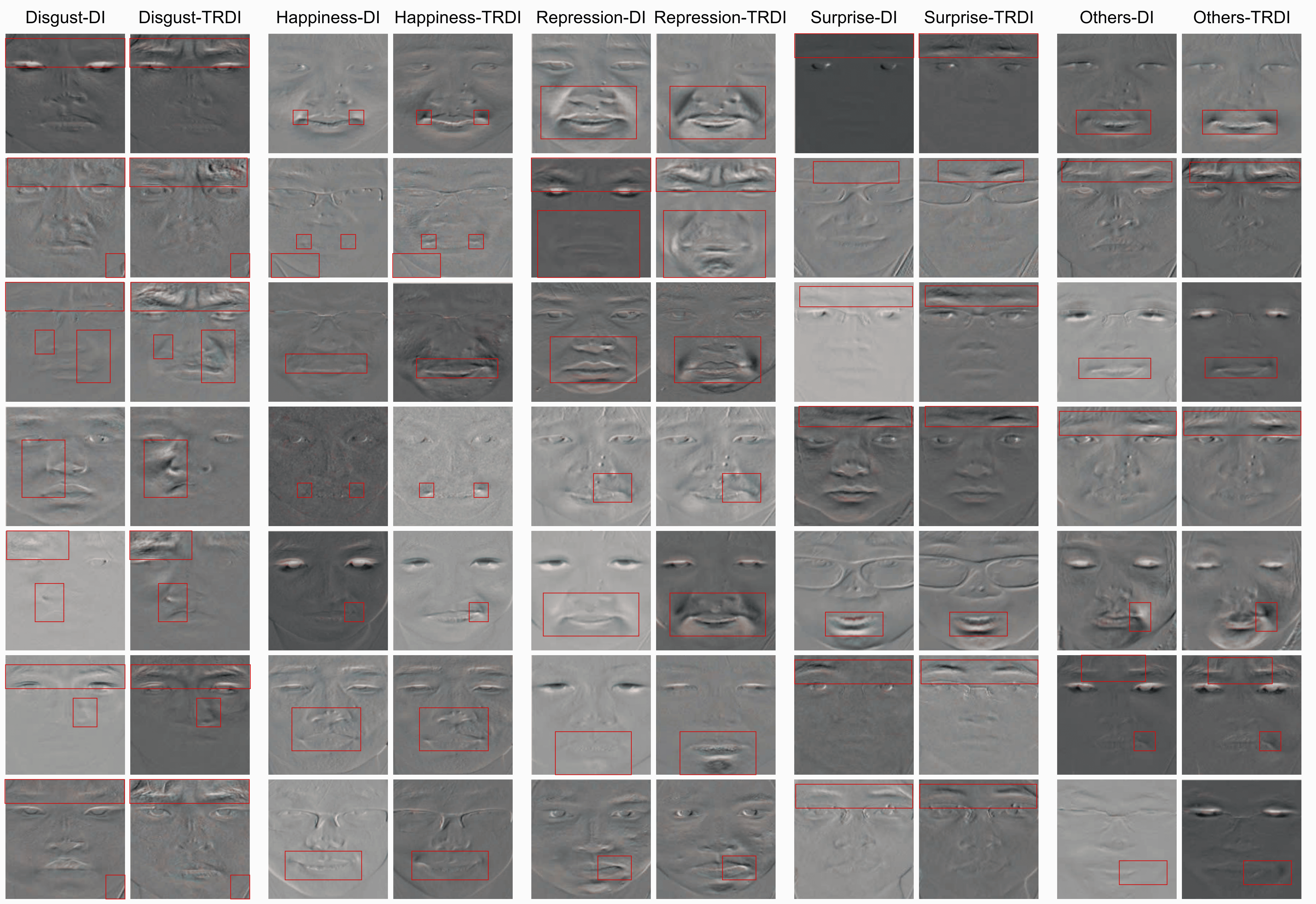}}
    \caption{The visualization of DIs and TRDIs}
    \label{fig:di12_visual}
\end{figure}

\textit{1) Visual representation}
The figure \ref{fig:di12_visual} illustrates the visualization of DIs and TRDIs extracted from facial video sequences. The DIs on the left represent the aggregated motion information over time, while the TRDIs on the right emphasize the residual features corresponding to frame-to-frame changes. Compared with DIs, TRDIs highlight more detailed facial dynamics such as wrinkles, micro muscle contractions, and fine motion cues around key regions of the face.

For instance, in Disgust, TRDIs show clearer activations around the nose and upper lip area, corresponding to AU9 (nose wrinkler) and AU10 (upper lip raiser). In Happiness, the eye corners and cheek areas become more pronounced, which are related to AU6 (cheek raiser) and AU12 (lip corner puller). Similarly, Repression and Surprise expressions show distinct movements around the mouth and eyebrows, emphasizing AU1 (inner brow raiser), AU2 (outer brow raiser), and AU5 (upper lid raiser) that are crucial for distinguishing subtle expressions.

Overall, TRDIs effectively preserve these fine-grained and transient expression cues, providing a richer representation of subtle facial motion patterns. This enhanced visualization confirms that TRDIs are better suited for capturing micro-expressions, where small and short-lived facial actions play a vital role in classification.

\begin{figure}
    \centering    \centerline{\includegraphics[width=1\linewidth]{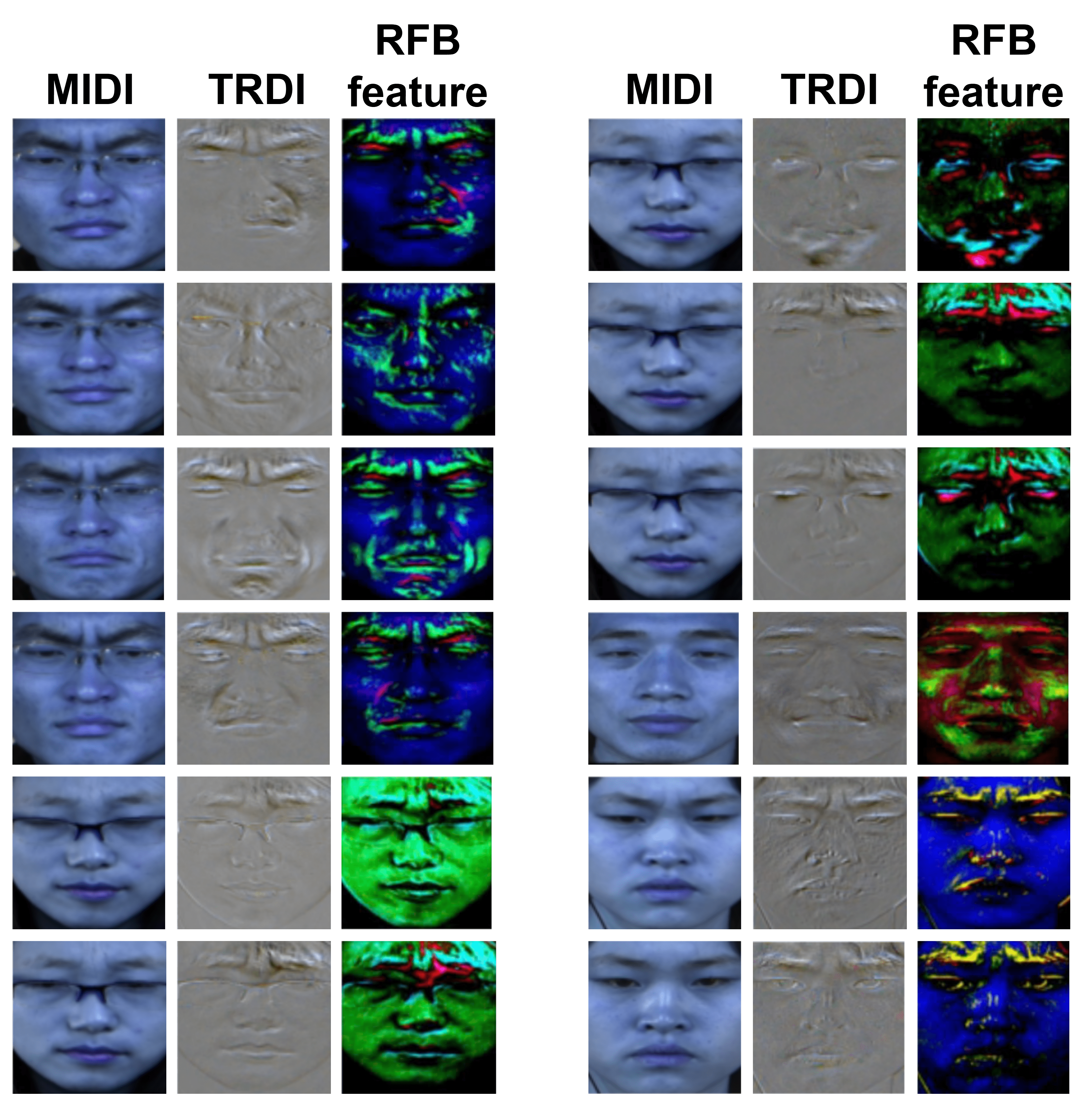}}
    \caption{The visualization of MIDI, TRDI, and feature}
    \label{fig:visual}
\end{figure}

This figure \ref{fig:visual} presents the visualization of MIDI inputs, TRDI, and the fused features extracted from the RFB. The first column shows MIDI, which captures global facial movements; the second column illustrates TRDI, which focuses on local variation information; and the third column displays the post-fusion features, which highlight the most important motion patterns for classification. The differences among the three representations demonstrate that the feature extraction and fusion process enables the model to learn both global and fine-grained dynamic information.

\textit{1) Experiment 1 result}

\begin{table*}[t]
    \centering
    \caption{Comparison of MER performance in terms of Accuracy (\%) on the CASME-II, SAMM and MMEW datasets between the different single input modalities and models}
    \resizebox{\textwidth}{!}{%
    \begin{tabular}{cccccccc}
        \toprule
        \multirow{2}{*}{No} & \multirow{2}{*}{Input modalities} &  
        \multicolumn{2}{c}{CASME-II} & \multicolumn{2}{c}{SAMM} & \multicolumn{2}{c}{MMEW} \\
        \cmidrule(lr){3-4}\cmidrule(lr){5-6}\cmidrule(lr){7-8}
        & & MSCAB(\%) & ResNet(\%) & MSCAB(\%) & ResNet(\%) & MSCAB(\%) & ResNet(\%)\\
        \midrule
        1 & Apex frame & 63.57 & 58.39 & 64.88 & 60.19 & 56.77 & 60.99 \\
        2 & Dynamic image & 63.51 & 65.29 & 60.78 & 62.02 & 56.9 & 54.9 \\
        3 & \textbf{MIDI (ours)} & 68.7 & 65.57 & 74.59 & 60.31 & 60.21 & 62.65 \\
        4 & \textbf{TRDI (ours)} & \textbf{81.06} & 76.57 & \textbf{71.85} & 66.4 & \textbf{68.58} & 66.78 \\
        \bottomrule
    \end{tabular}%
    }
    \label{tab:res3}
\end{table*}

Table \ref{tab:res3} present the experimental results comparing the performance of MER on the CASME-II, SAMM and MMEW datasets using different input modalities and backbones in two classify models: MSCAB (ours) and ResNet. The results clearly show that both TRDI and MIDI significantly improve accuracy compared to traditional methods such as Apex frame and dynamic image. On CASME-II, MSCAB with TRDI achieves 81.06\%, the highest among all cases, surpassing ResNet by 17.5\%; MIDI also brings a substantial improvement with 68.7\%. On SAMM, MIDI achieves the highest performance (74.59\%) with MSCAB, while TRDI achieves 71.85\%, both outperforming ResNet. On MMEW, TRDI continues to demonstrate superiority with 68.58\%, while MIDI achieves 60.21\%, both higher than ResNet. Notably, ResNet performs worse than MSCAB in most cases, suggesting that using a generic deep CNN model for MER is not necessarily the optimal choice. These results confirm that TRDI and MIDI, when combined with the classification model in the AFN architecture, provide superior and stable performance across diverse datasets.

\textit{2) Experiment 2 result}

\begin{table*}[t]
    \centering
    \caption{MER performance comparison with combined input modalities across CASME-II, SAMM, and MMEW datasets with different backbones in AFN architecture}
    \resizebox{\textwidth}{!}{%
    \begin{tabular}{cccccccc}
        \toprule
        \multirow{2}{*}{No} & \multirow{2}{*}{Input modalities} &  
        \multicolumn{2}{c}{CASME-II} & \multicolumn{2}{c}{SAMM} & \multicolumn{2}{c}{MMEW} \\
        \cmidrule(lr){3-4}\cmidrule(lr){5-6}\cmidrule(lr){7-8}
        & & MSCAB(\%) & ResNet(\%) & MSCAB(\%) & ResNet(\%) & MSCAB(\%) & ResNet(\%)\\
        \midrule
        1 & MIDI (ours) + DI & 62.65 &	63.54 &	63.35 &	60.16 &	61.72 &	58.7 \\
        2 & \textbf{MIDI (ours) + TRDI (ours)} & \textbf{93.95} & 85.27 & \textbf{80.16} & 74.36 & \textbf{76} & 71.68 \\
        \bottomrule
    \end{tabular}%
    }
    \label{tab:res4}
\end{table*}
The results in Table \ref{tab:res4} clearly show that combining modality approaches provides significant improvements over using each modality individually. In particular, the MIDI + TRDI combination achieves superior performance across all datasets. On CASME-II, MSCAB with MIDI + TRDI achieves 93.95\%, significantly higher than MIDI + DI (62.65\%) and surpassing the results reported in Experiment 1. On SAMM, this combination also attains the highest performance at 80.16\%, outperforming MIDI + DI by 16\%. On MMEW, the MIDI + TRDI combination achieves 76\%, which is 14.28\ higher than MIDI + DI. This demonstrates that TRDI and MIDI provide complementary information, which, when combined, creates a more effective dynamic representation for micro-expression recognition. Furthermore, in all cases, MSCAB outperforms ResNet, especially in the MIDI + TRDI combination, suggesting that a MER-specific backbone architecture such as MSCAB can exploit the complex features of the data more effectively. These results confirm that integrating multiple dynamic modalities, especially MIDI and TRDI, with a suitable classification model such as MSCAB within the AFN architecture is a powerful approach to improving the performance of MER.
\subsubsection{Comparison with State-of-the-Art (SOTA) Methods}

Table \ref{tab:performance} presents the performance comparison of MER methods on three datasets: CASME-II, SAMM, and MMEW, based on two main metrics: Accuracy (Acc) and Unweighted F1-score (UF1). In this comparison, the MMEW dataset uses six classes, excluding the Others label.

On CASME-II, the proposed method MIDI + TRDI + AFN achieves 93.95\% Accuracy and 89.72\% UF1, outperforming all other SOTA methods. In particular, the MMNet method – one of the recent strong baselines – achieves 88.35\% Accuracy and 86.76\% UF1, which are 5.17\% and 2.71\% lower than the proposed method, respectively. Other methods such as TSMicro and AUGCN yield significantly lower results, further demonstrating the effectiveness of our proposed architecture in extracting fine-grained features from ME data.

For SAMM, the proposed method achieves 82.47\% Accuracy and 66.52\% UF1. In terms of Accuracy, this result is higher than most other methods, only slightly behind DSSTDN with 85.68\% and MMNet with 85.21\%. However, in terms of UF1 – an important metric reflecting the balance across classes – our method achieves 66.52\%, which is higher than many SOTA methods such as AMAN (65.82\%) and MERsiamC3D (64.00\%). This indicates that the proposed method not only achieves high accuracy but also improves the ability to recognize expressions more evenly across different classes.

For the MMEW dataset, which includes six primary emotion categories, the proposed method also achieves impressive performance with 74.31\% Accuracy and 55.79\% UF1. Although these values are slightly lower than those obtained on CASME-II and SAMM, they are still higher than most other state-of-the-art (SOTA) methods listed in the table. This indicates that the model maintains stable recognition performance even under challenging conditions with high inter-class variation and complex facial dynamics.

These results demonstrate that the combination of MIDI, TRDI, and AFN modules effectively captures subtle and diverse facial movements, while enabling the model to generalize well across different datasets. The consistent and superior performance across multiple benchmarks confirms the robustness, adaptability, and reliability of the proposed method.
\begin{table}[h]
    \centering
    \caption{Comparison of MER performance in terms of Accuracy (\%) on the CASME-II, SAMM and MMEW datasets between the SOTA and our proposed method}
    \small
    \begin{tabular}{cp{5cm}cccccc}
        \toprule
        \multirow{2}{*}{No} & \multirow{2}{*}{Methods} &  \multicolumn{2}{c}{CASME-II} & \multicolumn{2}{c}{SAMM} & \multicolumn{2}{c}{MMEW} \\
        
        \\ \cmidrule(lr){3-4} \cmidrule(lr){4-4} \cmidrule(lr){5-6} \cmidrule(lr){6-6}\cmidrule(lr){7-8} \cmidrule(lr){8-8} & & Acc(\%) & UF1 & Acc(\%) & UF1& Acc(\%) & UF1\\
        \midrule
        \hline
        1 &  SODA4MER \cite{11095136}(2025) & 84.18 &  0.8141 & 80.30 &  0.7893& - & - \\
        2 &  AUGCN \cite{10446702}(2024) & 81.85 &  0.7760 & 79.82 & 0.7568& - & -  \\
        3 &  KPCANet \cite{10.1145/3607829.3616444}(2023) & 70.46 & 0.6592 & 63.83 & 0.5215& - & -  \\
        4 &  MMNet \cite{li2022mmnetmusclemotionguidednetwork}(2022) & 88.35 & 0.8676 & 80.14 &  0.7291& - & -  \\
        5 &  KFC-MER \cite{9428407}(2021) & 72.76 &  0.7375 & 63.24 & 0.5709& - & -  \\
        6 &  Graph-TCN \cite{10.1145/3394171.3413714}(2020) & 73.98 &  0.7246 & 75.00 &  0.6985& - & -  \\
        7 &  TSFmicro \cite{unknown}(2025) & 87.50 &  86.17 & 80.88 &  77.00& - & -  \\
        8 &  DSSTDN \cite{Matharaarachchi2025}(2025) & 77.91 &  77.55 & 78.68 & 76.41& - & -  \\
        9 &  C3DBed \cite{PAN2023106258}(2023) & 77.64 & 75.20 & 75.73 &  72.16 & - & - \\
        10 &  AMAN \cite{9747232}(2022) & 75.40 & 71.25 & 68.85 &66.82& - & -  \\
        11 &  MTMNet \cite{li2022mmnetmusclemotionguidednetwork}(2022) & 86.09 & 85.21 & 80.14 & 72.91& - & -  \\
        12 &  MERSiamC3D \cite{ZHAO2021276}(2021) & 81.89 & 83.00 & 68.75 & 64.00& - & -  \\
        13 &  MOL \cite{11045217}(2025) & 79.23 & - & 76.68 & - & - & - \\
        14 &   Multi-task mid-level feature learning\cite{HE201744}(2017) & - & - &  55.0 & -  & 54.2&- \\
        15 &  KGSL\cite{8328869}(2018) & - & - & 48.6 & - & 56.9&- \\
        16 & SparseMDMO\cite{8408485}(2019) & - & - & 52.9  & - & 60&- \\
        17 &  MDMO\cite{7286757}(2016)& - & - &  50 & - &65.7&- \\
        18 &  DTSCNN\cite{dual}(2017) & - & - & 69.2 & - &65.9&- \\
        19 & TLCNN\cite{WANG2018251}(2018) & - & - & 73.5 & - &69.4 & - \\
        20 & \textbf{MIDI + TRDI + AFN (ours)} & \textbf{93.95} & \textbf{89.72} & \textbf{80.47} & \textbf{66.52} & \textbf{73.41} &  \textbf{55.79}\\
        \bottomrule
    \end{tabular}
    \label{tab:performance}
\end{table}
\subsection{Ablation study}

To further verify the validity and effectiveness of the proposed method, we design three ablation studies to address the following questions:
\begin{itemize}
    \item \textit{Is the RFB really necessary?}
        
        We compare it with a manual fusion strategy to verify whether the original design provides a clear advantage.
    \item \textit{Are the components within the RFB important?}
        
        We remove or fix each component individually to analyze their contribution and evaluate their impact on the overall performance.
    \item \textit{Do the component removal and fixation strategies work similarly on the MER model?}
        
        We apply the same ablation procedure to the MER model to examine whether the trend remains consistent, thereby confirming the generality of the proposed framework.
\end{itemize}

\subsubsection{Necessity of RFB}
In this section, we design a manual fusion strategy for two inputs, TRDI and MIDI, based on the formula as shown in Equation \ref{equa11}:
\begin{equation}
\text{FusedFeature} = \lambda \cdot \text{TRDI} + (1-\lambda) \cdot \text{MIDI}
\label{equa11}
\end{equation}

where $\lambda \in [0,1]$ is a coefficient that adjusts the contribution of each feature to the fused representation before being fed into the \textit{MSCAB} for classification.

We choose this linear formulation because of its simplicity, intuitiveness, and ease of control. It allows direct verification of the relative contributions of TRDI and MIDI to the fused feature and provides a reasonable baseline for comparison with the RFB. Through this, we can assess whether the original design of the fusion module offers a superior advantage.

To verify the necessity of the RFB, we conduct two experimental steps.

First, we design a manual fusion strategy by varying the coefficient $\alpha \in [0,1]$ on three datasets, CASME II, SAMM, and MEWW, following the LOSO protocol. The results in Table~\ref{tab:manual_fusion} show that the highest performance is achieved when $\alpha$ is in the range of $0.5 \sim 0.6$. This indicates that TRDI and MIDI contribute almost equally, and that a balanced combination of these two feature sources yields the optimal performance.

\begin{table}[h]
    \centering
    \caption{MER performance (Accuracy \%) on the CASME-II, SAMM, and MMEW datasets under the LOSO setting with different $\lambda$ values}
    \label{tab:manual_fusion}
    \begin{tabular}{c ccc}
        \toprule
        \multirow{2}{*}{$\lambda$} & \multicolumn{3}{c}{LOSO -- MSCAB} \\ 
        \cmidrule(lr){2-4}
        & CASME & SAMM & MMEW \\
        \midrule
        0   & 68.7   & 74.59 & 60.21 \\
        0.1 & 76.13  & 75    & 63.48 \\
        0.2 & 81.53  & 74.38 & 61.37 \\
        0.3 & 87.50  & 75.74 & 65.55 \\
        0.4 & 85.16  & 74.88 & 63.13 \\
        0.5 & 88.20  & 73.25 & 72.69 \\
        0.6 & 87.79  & 76.96 & 68.19 \\
        0.7 & 86.89  & 75.25 & 67.70 \\
        0.8 & 87.97  & 73.35 & 67.66 \\
        0.9 & 87.18  & 74.51 & 69.73 \\
        1.0 & 81.06  & 71.85 & 68.58 \\
        \bottomrule
    \end{tabular}
\end{table}

Next, we compare the results of the manual fusion strategy with those of the proposed RFB. To ensure objectivity, we also include ResNet as a baseline, trained on the best fusion features. The results in Table~\ref{tab:comparison_fusion} show that although manual fusion achieved competitive results, the RFB consistently outperformed it on all three datasets. In particular, MSCAB consistently surpassed ResNet in every case, demonstrating that MSCAB is more effective at exploiting dynamic information and is better suited for the micro-expression recognition task.

\begin{table}[h]
    \centering
    \caption{Comparison of manual fusion, RFB (ours), and ResNet baseline in terms of Accuracy (\%) on the CASME-II, SAMM, and MMEW datasets}
    \resizebox{\textwidth}{!}{%
    \begin{tabular}{cccccccc}
        \toprule
        \multirow{2}{*}{No} & \multirow{2}{*}{Input modalities} &  
        \multicolumn{2}{c}{CASME-II} & \multicolumn{2}{c}{SAMM} & \multicolumn{2}{c}{MMEW} \\
        \cmidrule(lr){3-4}\cmidrule(lr){5-6}\cmidrule(lr){7-8}
        & & MSCAB(\%) & ResNet(\%) & MSCAB(\%) & ResNet(\%) & 
        MSCAB(\%) & ResNet(\%)\\
        \midrule
        1 & FusedFeature & 88.2 &	80.55 &	76.96 &	68.33 &	72.69 &	70.93 \\
        2 & MIDI + TRDI (ours) & 93.95 &	85.27 &	80.16 &	74.36 &	76 &	71.68 \\
        \bottomrule
    \end{tabular}%
    }
    \label{tab:comparision_fusion}
\end{table}

In summary, these results demonstrate that the RFB is not only necessary but also more effective than both manual fusion and the baseline ResNet, due to its ability to automatically learn complex interactions between TRDI and MIDI rather than relying solely on fixed linear combinations.
\subsubsection{Effect of RFB components}

To evaluate the contribution of each component in the RFB in detail, we conducted an ablation experiment by removing the residual connection and the attention mechanism individually. The results are presented in Table~\ref{tab:ablation_fusion}.
\begin{table}[h!]
\centering
\caption{Performance comparison under different component configurations of RFB.}
\begin{tabular}{lccc}
\hline
\textbf{Setting} & \textbf{CASME II} & \textbf{SAMM} & \textbf{MEWW} \\
\hline
No-attn + residual      & 76.56 & 68.03 & 66.93 \\
Attn + no-residual      & 76.71 & 66.56 & 67.19 \\
No-attn + no-residual   & 70.09 & 57.96 & 60.41 \\
Full                    & 93.95 & 80.16 & 75.99 \\
\hline
\end{tabular}
\label{tab:ablation_fusion}
\end{table}

When only the residual connection (\textit{No-attn + residual}) is retained, the performance is significantly degraded compared to the full model 
(CASME-II: 76.56\%, SAMM: 68.03\%, MMEW: 66.93\%).

When only the attention mechanism (\textit{Attn + no-residual}) is retained, the accuracy slightly improves compared to the previous case but is still lower than the full model 
(CASME-II: 76.71\%, SAMM: 66.56\%, MMEW: 67.19\%).

When both components are removed (\textit{No-attn + no-residual}), the model suffers severe performance degradation across all three datasets 
(CASME-II: 70.09\%, SAMM: 57.96\%, MMEW: 60.41\%).

In contrast, the \textit{Full} model with both residual connection and attention mechanism achieves the best results 
(CASME-II: 93.95\%, SAMM: 80.16\%, MMEW: 75.99\%), 
demonstrating the effectiveness of combining both components simultaneously.

These results confirm that both the residual connection and the attention mechanism play essential roles in enhancing the representational capacity of the RFB, and their combination brings a significant improvement in MER performance.

\subsubsection{Effect of MSCAB module components}

To evaluate the role of the components in \textit{MSCAB} in detail, we conducted a series of experiments by removing or fixing each component in turn, including the learned weights $\alpha$ and the SE block. The results are presented in Table~\ref{tab:ablation_MSCAB}.

\begin{table}[h!]
\centering
\caption{Performance comparison under different component configurations of MSCAB}
\begin{tabular}{lccc}
\hline
\textbf{Setting} & \textbf{CASME II} & \textbf{SAMM} & \textbf{MEWW} \\
\hline
$\alpha = \tfrac{1}{4}$ & 72.51 & 63.85 & 62.56 \\	
no SE & 75.56 & 63.97 & 63.69 \\
$\alpha = \tfrac{1}{4}$ + no SE & 73.36 & 51.25 & 58.56 \\
Full & 93.95 & 80.16 & 76.00 \\
\hline
\end{tabular}
\label{tab:ablation_MSCAB}
\end{table}

When $\alpha$ is fixed to a uniform value of $1/4$ for each branch (\textit{a\_fixed}), the performance drops significantly compared to the full model (CASME-II: 72.51\%, SAMM: 63.85\%, MMEW: 62.56\%). 

When the SE block is removed (\textit{no se}), the model slightly improves compared to the \textit{a\_fixed} case but still falls short of the full model (CASME-II: 75.56\%, SAMM: 63.97\%, MMEW: 63.69\%). 

When simultaneously fixing $\alpha$ and removing the SE block (\textit{a\_fixed + no se}), the model’s performance deteriorates drastically, especially on the SAMM dataset (CASME-II: 73.36\%, SAMM: 51.25\%, MMEW: 58.56\%). 

In contrast, the full model with both the learned $\alpha$ weights and the SE block achieves the best performance across all three datasets (CASME-II: 93.95\%, SAMM: 80.16\%, MMEW: 75.99\%). This confirms that both components play an essential role, and combining them optimizes the representational capacity, thereby significantly improving the MER performance.

\section{Conclusion}
\label{s:conclusion}
This paper presents a novel MER method that combines two specially designed dynamic image representations – TRDI and MIDI – with an AFN. Unlike traditional dynamic images, TRDI emphasizes the temporal progression around the apex frame, while MIDI highlights subtle motion intensity cues. By adaptively fusing these two complementary representations via AFN, our method enhances discriminative feature learning while reducing irrelevant noise.

Experiments on three benchmark datasets (CASME-II, SAMM, and MMEW) demonstrate the superiority of the proposed method. In particular, the combination of TRDI and MIDI with AFN achieves 93.95\% Accuracy and 89.72\% UF1 on CASME-II, setting a new state-of-the-art benchmark. On SAMM and MMEW, the method also delivers competitive results, with significant improvements in UF1, confirming more balanced recognition across expression classes. The ablation studies further demonstrate the necessity of the RFB and the components of MSCAB, thereby emphasizing the effectiveness of the adaptive fusion mechanism and attention.

Overall, the proposed framework not only improves the performance of MER but also provides a robust solution to the challenges of subtle intensity, short duration, and limited data in MEs. In the future, this research can be extended by integrating additional multi-modal signals such as physiological data, and applied to real-world scenarios such as lie detection, affective computing, and human-computer interaction. Ultimately, our contribution not only advances the development of the MER field but also establishes a solid foundation for further innovations and practical applications in affective computing.







\bibliographystyle{elsarticle-num} 
\bibliography{bibliography}

\begin{thebibliography}{10}
\expandafter\ifx\csname url\endcsname\relax
  \def\url#1{\texttt{#1}}\fi
\expandafter\ifx\csname urlprefix\endcsname\relax\def\urlprefix{URL }\fi
\expandafter\ifx\csname href\endcsname\relax
  \def\href#1#2{#2} \def\path#1{#1}\fi

\bibitem{10144523}
G.~Zhao, X.~Li, Y.~Li, M.~Pietikäinen, Facial micro-expressions: An overview, Proceedings of the IEEE 111~(10) (2023) 1215--1235.
\newblock \href {https://doi.org/10.1109/JPROC.2023.3275192} {\path{doi:10.1109/JPROC.2023.3275192}}.

\bibitem{mehrabian1967decoding}
A.~Mehrabian, M.~Wiener, Decoding of inconsistent communications, Journal of Personality and Social Psychology 6~(1) (1967) 109.

\bibitem{amsel2019urban}
T.~T. Amsel, An Urban Legend Called: ‘The 7/38/55 Ratio Rule’, 2nd Edition, Vol.~13, Sciendo, De Gruyter Poland Sp. z o.o., Warsaw, Poland, 2019.

\bibitem{Bhushan2015}
B.~Bhushan, Study of Facial Micro-expressions in Psychology, Springer India, New Delhi, 2015, pp. 265--286.
\newblock \href {https://doi.org/10.1007/978-81-322-1934-7\_13} {\path{doi:10.1007/978-81-322-1934-7\_13}}.

\bibitem{yan2013fast}
W.-J. Yan, Q.~Wu, J.~Liang, Y.-H. Chen, X.~Fu, How fast are the leaked facial expressions: The duration of micro-expressions, Journal of Nonverbal Behavior 37 (2013) 217--230.

\bibitem{polikovsky2010detection}
S.~Polikovsky, Y.~Kameda, Y.~Ohta, Detection and measurement of facial micro-expression characteristics for psychological analysis, Kameda’s Publication 110 (2010) 57--64.

\bibitem{li2022deeplearningmicroexpressionrecognition}
Y.~Li, J.~Wei, Y.~Liu, J.~Kauttonen, G.~Zhao, \href{https://arxiv.org/abs/2107.02823}{Deep learning for micro-expression recognition: A survey} (2022).
\newblock \href {http://arxiv.org/abs/2107.02823} {\path{arXiv:2107.02823}}.
\newline\urlprefix\url{https://arxiv.org/abs/2107.02823}

\bibitem{bilen2017actionrecognitiondynamicimage}
H.~Bilen, B.~Fernando, E.~Gavves, A.~Vedaldi, \href{https://arxiv.org/abs/1612.00738}{Action recognition with dynamic image networks} (2017).
\newblock \href {http://arxiv.org/abs/1612.00738} {\path{arXiv:1612.00738}}.
\newline\urlprefix\url{https://arxiv.org/abs/1612.00738}

\bibitem{9207718}
M.~Verma, S.~K. Vipparthi, G.~Singh, Non-linearities improve originet based on active imaging for micro expression recognition, in: 2020 International Joint Conference on Neural Networks (IJCNN), 2020, pp. 1--8.
\newblock \href {https://doi.org/10.1109/IJCNN48605.2020.9207718} {\path{doi:10.1109/IJCNN48605.2020.9207718}}.

\bibitem{verma2021affectivenetaffectivemotionfeaturelearningfor}
M.~Verma, S.~K. Vipparthi, G.~Singh, \href{https://arxiv.org/abs/2104.07569}{Affectivenet: Affective-motion feature learningfor micro expression recognition} (2021).
\newblock \href {http://arxiv.org/abs/2104.07569} {\path{arXiv:2104.07569}}.
\newline\urlprefix\url{https://arxiv.org/abs/2104.07569}

\bibitem{zhao2007dynamic}
G.~Zhao, M.~Pietikäinen, Dynamic texture recognition using local binary patterns with an application to facial expressions, IEEE Transactions on Pattern Analysis and Machine Intelligence (2007).

\bibitem{10.1007/978-3-319-16865-4_34}
Y.~Wang, J.~See, R.~C.-W. Phan, Y.-H. Oh, Lbp with six intersection points: Reducing redundant information in lbp-top for micro-expression recognition, in: D.~Cremers, I.~Reid, H.~Saito, M.-H. Yang (Eds.), Computer Vision -- ACCV 2014, Springer International Publishing, Cham, 2015, pp. 525--537.

\bibitem{Wang2015EfficientSL}
Y.~Wang, J.~See, R.~C.-W. Phan, Y.-H. Oh, \href{https://api.semanticscholar.org/CorpusID:16115916}{Efficient spatio-temporal local binary patterns for spontaneous facial micro-expression recognition}, PLoS ONE 10 (2015).
\newline\urlprefix\url{https://api.semanticscholar.org/CorpusID:16115916}

\bibitem{Li2016SpontaneousFM}
X.~Li, J.~Yu, S.~Zhan, \href{https://api.semanticscholar.org/CorpusID:12781885}{Spontaneous facial micro-expression detection based on deep learning}, 2016 IEEE 13th International Conference on Signal Processing (ICSP) (2016) 1130--1134.
\newline\urlprefix\url{https://api.semanticscholar.org/CorpusID:12781885}

\bibitem{Liu2022}
Y.~Liu, Y.~Li, X.~Yi, Z.~Hu, H.~Zhang, Y.~Liu, \href{https://doi.org/10.1038/s41598-022-21738-8}{Micro-expression recognition model based on tv-l1 optical flow method and improved shufflenet}, Scientific Reports 12~(1) (2022) 17522.
\newblock \href {https://doi.org/10.1038/s41598-022-21738-8} {\path{doi:10.1038/s41598-022-21738-8}}.
\newline\urlprefix\url{https://doi.org/10.1038/s41598-022-21738-8}

\bibitem{9567872}
A.~F. Ibrahim, S.~P. Ristiawanto, C.~Setianingsih, B.~Irawan, Micro-expression recognition using vgg19 convolutional neural network architecture and random forest, in: 2021 4th International Symposium on Agents, Multi-Agent Systems and Robotics (ISAMSR), 2021, pp. 150--156.
\newblock \href {https://doi.org/10.1109/ISAMSR53229.2021.9567872} {\path{doi:10.1109/ISAMSR53229.2021.9567872}}.

\bibitem{10248754}
L.~Wang, S.~Zhan, Face micro-expression recognition algorithm based on resnet depth model, in: 2023 8th International Conference on Intelligent Computing and Signal Processing (ICSP), 2023, pp. 1533--1537.
\newblock \href {https://doi.org/10.1109/ICSP58490.2023.10248754} {\path{doi:10.1109/ICSP58490.2023.10248754}}.

\bibitem{Zhou2023}
H.~Zhou, S.~Huang, Y.~Xu, \href{https://doi.org/10.1007/s00530-023-01164-0}{Inceptr: micro-expression recognition integrating inception-cbam and vision transformer}, Multimedia Systems 29~(6) (2023) 3863--3876.
\newblock \href {https://doi.org/10.1007/s00530-023-01164-0} {\path{doi:10.1007/s00530-023-01164-0}}.
\newline\urlprefix\url{https://doi.org/10.1007/s00530-023-01164-0}

\bibitem{inproceedings5}
M.~Bai, R.~Goecke, Investigating lstm for micro-expression recognition, 2020, pp. 7--11.
\newblock \href {https://doi.org/10.1145/3395035.3425248} {\path{doi:10.1145/3395035.3425248}}.

\bibitem{Li2019}
J.~Li, Y.~Wang, J.~See, W.~Liu, \href{https://doi.org/10.1007/s10044-018-0757-5}{Micro-expression recognition based on 3d flow convolutional neural network}, Pattern Analysis and Applications 22~(4) (2019) 1331--1339.
\newblock \href {https://doi.org/10.1007/s10044-018-0757-5} {\path{doi:10.1007/s10044-018-0757-5}}.
\newline\urlprefix\url{https://doi.org/10.1007/s10044-018-0757-5}

\bibitem{sanchezcaballero20203dfcnnrealtimeactionrecognition}
A.~Sanchez-Caballero, S.~de~López-Diz, D.~Fuentes-Jimenez, C.~Losada-Gutiérrez, M.~Marrón-Romera, D.~Casillas-Perez, M.~I. Sarker, \href{https://arxiv.org/abs/2006.07743}{3dfcnn: Real-time action recognition using 3d deep neural networks with raw depth information} (2020).
\newblock \href {http://arxiv.org/abs/2006.07743} {\path{arXiv:2006.07743}}.
\newline\urlprefix\url{https://arxiv.org/abs/2006.07743}

\bibitem{inproceedings1}
L.~Qing, F.~Huang, Micro-expression recognition based on capsule network, 2024, pp. 871--878.
\newblock \href {https://doi.org/10.1109/CSIS-IAC63491.2024.10919257} {\path{doi:10.1109/CSIS-IAC63491.2024.10919257}}.

\bibitem{10.1007/978-981-96-1071-6_4}
Y.~Guo, T.~Xie, W.~Jia, S.~Xu, X.~Ben, Micro-expression recognition via cnn and multi-path vision transformer integrated with spatiotemporal separated self-attention, in: S.~Yu, W.~Jia, X.~Shu, X.~Yuan, J.~Gui, J.~Tang, C.~Shan, Q.~Liu (Eds.), Biometric Recognition, Springer Nature Singapore, Singapore, 2025, pp. 35--46.

\bibitem{inproceedings10}
M.~Bai, R.~Goecke, Investigating lstm for micro-expression recognition, 2020, pp. 7--11.
\newblock \href {https://doi.org/10.1145/3395035.3425248} {\path{doi:10.1145/3395035.3425248}}.

\bibitem{3dresnet}
H.~Jin, N.~He, Z.~Li, P.~Yang, \href{https://www.aimspress.com/article/doi/10.3934/mbe.2024221}{Micro-expression recognition based on multi-scale 3d residual convolutional neural network}, Mathematical Biosciences and Engineering 21~(4) (2024) 5007--5031.
\newblock \href {https://doi.org/10.3934/mbe.2024221} {\path{doi:10.3934/mbe.2024221}}.
\newline\urlprefix\url{https://www.aimspress.com/article/doi/10.3934/mbe.2024221}

\bibitem{fernando2015modeling}
B.~Fernando, E.~Gavves, J.~M. Oramas, A.~Ghodrati, T.~Tuytelaars, Modeling video evolution for action recognition, in: Proceedings of the IEEE conference on computer vision and pattern recognition, 2015, pp. 5378--5387.

\bibitem{fernando2016rank}
B.~Fernando, E.~Gavves, J.~Oramas, A.~Ghodrati, T.~Tuytelaars, Rank pooling for action recognition, IEEE transactions on pattern analysis and machine intelligence 39~(4) (2016) 773--787.

\bibitem{Yan2014-za}
W.-J. Yan, X.~Li, S.-J. Wang, G.~Zhao, Y.-J. Liu, Y.-H. Chen, X.~Fu, {CASME} {II}: an improved spontaneous micro-expression database and the baseline evaluation, PLoS One 9~(1) (2014) e86041.

\bibitem{7492264}
A.~K. Davison, C.~Lansley, N.~Costen, K.~Tan, M.~H. Yap, Samm: A spontaneous micro-facial movement dataset, IEEE Transactions on Affective Computing 9~(1) (2018) 116--129.
\newblock \href {https://doi.org/10.1109/TAFFC.2016.2573832} {\path{doi:10.1109/TAFFC.2016.2573832}}.

\bibitem{9382112}
X.~Ben, Y.~Ren, J.~Zhang, S.-J. Wang, K.~Kpalma, W.~Meng, Y.-J. Liu, Video-based facial micro-expression analysis: A survey of datasets, features and algorithms, IEEE Transactions on Pattern Analysis and Machine Intelligence 44~(9) (2022) 5826--5846.
\newblock \href {https://doi.org/10.1109/TPAMI.2021.3067464} {\path{doi:10.1109/TPAMI.2021.3067464}}.

\bibitem{11095136}
B.~Zhang, X.~Wang, C.~Wang, G.~He, Dynamic stereotype theory induced micro-expression recognition with oriented deformation, in: 2025 IEEE/CVF Conference on Computer Vision and Pattern Recognition (CVPR), 2025, pp. 10701--10711.
\newblock \href {https://doi.org/10.1109/CVPR52734.2025.01000} {\path{doi:10.1109/CVPR52734.2025.01000}}.

\bibitem{10446702}
L.~Wang, P.~Huang, W.~Cai, X.~Liu, Micro-expression recognition by fusing action unit detection and spatio-temporal features, in: ICASSP 2024 - 2024 IEEE International Conference on Acoustics, Speech and Signal Processing (ICASSP), 2024, pp. 5595--5599.
\newblock \href {https://doi.org/10.1109/ICASSP48485.2024.10446702} {\path{doi:10.1109/ICASSP48485.2024.10446702}}.

\bibitem{10.1145/3607829.3616444}
W.~Feng, M.~Xu, Y.~Chen, X.~Wang, J.~Guo, L.~Dai, N.~Wang, X.~Zuo, X.~Li, \href{https://doi.org/10.1145/3607829.3616444}{Nonlinear deep subspace network for micro-expression recognition}, in: Proceedings of the 3rd Workshop on Facial Micro-Expression: Advanced Techniques for Multi-Modal Facial Expression Analysis, FME '23, Association for Computing Machinery, New York, NY, USA, 2023, p. 1–8.
\newblock \href {https://doi.org/10.1145/3607829.3616444} {\path{doi:10.1145/3607829.3616444}}.
\newline\urlprefix\url{https://doi.org/10.1145/3607829.3616444}

\bibitem{li2022mmnetmusclemotionguidednetwork}
H.~Li, M.~Sui, Z.~Zhu, F.~Zhao, \href{https://arxiv.org/abs/2201.05297}{Mmnet: Muscle motion-guided network for micro-expression recognition} (2022).
\newblock \href {http://arxiv.org/abs/2201.05297} {\path{arXiv:2201.05297}}.
\newline\urlprefix\url{https://arxiv.org/abs/2201.05297}

\bibitem{9428407}
Y.~Su, J.~Zhang, J.~Liu, G.~Zhai, Key facial components guided micro-expression recognition based on first second-order motion, in: 2021 IEEE International Conference on Multimedia and Expo (ICME), 2021, pp. 1--6.
\newblock \href {https://doi.org/10.1109/ICME51207.2021.9428407} {\path{doi:10.1109/ICME51207.2021.9428407}}.

\bibitem{10.1145/3394171.3413714}
L.~Lei, J.~Li, T.~Chen, S.~Li, \href{https://doi.org/10.1145/3394171.3413714}{A novel graph-tcn with a graph structured representation for micro-expression recognition}, in: Proceedings of the 28th ACM International Conference on Multimedia, MM '20, Association for Computing Machinery, New York, NY, USA, 2020, p. 2237–2245.
\newblock \href {https://doi.org/10.1145/3394171.3413714} {\path{doi:10.1145/3394171.3413714}}.
\newline\urlprefix\url{https://doi.org/10.1145/3394171.3413714}

\bibitem{unknown}
F.~Liu, B.~Nan, X.~Qian, X.~Fu, Temporal and spatial feature fusion framework for dynamic micro expression recognition (05 2025).
\newblock \href {https://doi.org/10.48550/arXiv.2505.16372} {\path{doi:10.48550/arXiv.2505.16372}}.

\bibitem{Matharaarachchi2025}
N.~Matharaarachchi, M.~F. Pasha, \href{https://doi.org/10.1007/s00521-024-10374-0}{A dual stream spatio-temporal deep network for micro-expression recognition using upper facial features}, Neural Computing and Applications 37~(3) (2025) 1271--1287.
\newblock \href {https://doi.org/10.1007/s00521-024-10374-0} {\path{doi:10.1007/s00521-024-10374-0}}.
\newline\urlprefix\url{https://doi.org/10.1007/s00521-024-10374-0}

\bibitem{PAN2023106258}
H.~Pan, L.~Xie, Z.~Wang, C3dbed: Facial micro-expression recognition with three-dimensional convolutional neural network embedding in transformer model, Engineering Applications of Artificial Intelligence 123 (2023) 106258.
\newblock \href {https://doi.org/https://doi.org/10.1016/j.engappai.2023.106258} {\path{doi:https://doi.org/10.1016/j.engappai.2023.106258}}.

\bibitem{9747232}
M.~Wei, W.~Zheng, Y.~Zong, X.~Jiang, C.~Lu, J.~Liu, A novel micro-expression recognition approach using attention-based magnification-adaptive networks, in: ICASSP 2022 - 2022 IEEE International Conference on Acoustics, Speech and Signal Processing (ICASSP), 2022, pp. 2420--2424.
\newblock \href {https://doi.org/10.1109/ICASSP43922.2022.9747232} {\path{doi:10.1109/ICASSP43922.2022.9747232}}.

\bibitem{ZHAO2021276}
S.~Zhao, H.~Tao, Y.~Zhang, T.~Xu, K.~Zhang, Z.~Hao, E.~Chen, A two-stage 3d cnn based learning method for spontaneous micro-expression recognition, Neurocomputing 448 (2021) 276--289.
\newblock \href {https://doi.org/https://doi.org/10.1016/j.neucom.2021.03.058} {\path{doi:https://doi.org/10.1016/j.neucom.2021.03.058}}.

\bibitem{11045217}
Z.~Shao, Y.~Cheng, F.~Li, Y.~Zhou, X.~Lu, Y.~Xie, L.~Ma, Mol: Joint estimation of micro-expression, optical flow, and landmark via transformer-graph-style convolution, IEEE Transactions on Pattern Analysis and Machine Intelligence (2025) 1--14\href {https://doi.org/10.1109/TPAMI.2025.3581162} {\path{doi:10.1109/TPAMI.2025.3581162}}.

\bibitem{HE201744}
J.~He, J.-F. Hu, X.~Lu, W.-S. Zheng, \href{https://www.sciencedirect.com/science/article/pii/S0031320316303879}{Multi-task mid-level feature learning for micro-expression recognition}, Pattern Recognition 66 (2017) 44--52.
\newblock \href {https://doi.org/https://doi.org/10.1016/j.patcog.2016.11.029} {\path{doi:https://doi.org/10.1016/j.patcog.2016.11.029}}.
\newline\urlprefix\url{https://www.sciencedirect.com/science/article/pii/S0031320316303879}

\bibitem{8328869}
Y.~Zong, X.~Huang, W.~Zheng, Z.~Cui, G.~Zhao, Learning from hierarchical spatiotemporal descriptors for micro-expression recognition, IEEE Transactions on Multimedia 20~(11) (2018) 3160--3172.
\newblock \href {https://doi.org/10.1109/TMM.2018.2820321} {\path{doi:10.1109/TMM.2018.2820321}}.

\bibitem{8408485}
Y.-J. Liu, B.-J. Li, Y.-K. Lai, Sparse mdmo: Learning a discriminative feature for micro-expression recognition, IEEE Transactions on Affective Computing 12~(1) (2021) 254--261.
\newblock \href {https://doi.org/10.1109/TAFFC.2018.2854166} {\path{doi:10.1109/TAFFC.2018.2854166}}.

\bibitem{7286757}
Y.-J. Liu, J.-K. Zhang, W.-J. Yan, S.-J. Wang, G.~Zhao, X.~Fu, A main directional mean optical flow feature for spontaneous micro-expression recognition, IEEE Transactions on Affective Computing 7~(4) (2016) 299--310.
\newblock \href {https://doi.org/10.1109/TAFFC.2015.2485205} {\path{doi:10.1109/TAFFC.2015.2485205}}.

\bibitem{dual}
M.~Peng, C.~Wang, T.~Chen, G.~Liu, X.~Fu, Dual temporal scale convolutional neural network for micro-expression recognition, Frontiers in psychology 8 (2017) 1745.

\bibitem{WANG2018251}
S.-J. Wang, B.-J. Li, Y.-J. Liu, W.-J. Yan, X.~Ou, X.~Huang, F.~Xu, X.~Fu, Micro-expression recognition with small sample size by transferring long-term convolutional neural network, Neurocomputing 312 (2018) 251--262.
\newblock \href {https://doi.org/https://doi.org/10.1016/j.neucom.2018.05.107} {\path{doi:https://doi.org/10.1016/j.neucom.2018.05.107}}.

\end{thebibliography}

\end{document}